\documentclass{article}

\usepackage[preprint]{corl_2026} 
\usepackage{amsmath}
\usepackage{amssymb}
\usepackage{booktabs}
\usepackage{tikz}
\usepackage{pgfplots}
\pgfplotsset{compat=1.18}
\usetikzlibrary{positioning,arrows.meta,shapes.geometric,fit,backgrounds,calc,decorations.pathreplacing}
\usepackage{multirow}
\title{Optimality-Preserving Decomposition for Scalable QAOA in Natural-Language-Guided Multi-Drone Assignment}

\author{
  Junyeop Bang\\
  Department of Smart Convergence\\
  Korea University 
  Republic of Korea\\
  \texttt{junyeop9981@korea.ac.kr} \\
  \And
  Byongho Lee\\
  Department of Smart Convergence\\
  Korea University 
  Republic of Korea\\
  \texttt{unlike96@korea.ac.kr} \\
  \And
  Dohyun An\\
  Department of Smart Convergence\\
  Korea University 
  Republic of Korea\\
  \texttt{sinsiga1905@korea.ac.kr} \\
  \And
  Hwangnam Kim\\
  School of Electrical Engineering\\
  Korea University 
  Republic of Korea\\
  \texttt{hnkim@korea.ac.kr} \\
}

\begin{document}
\maketitle

\begin{abstract}
As multi-drone fleets scale, zone assignment rapidly evolves into an intractable NP-hard combinatorial problem that overwhelms classical exhaustive search. While quantum optimization promises to shatter these classical bottlenecks, mapping complex spatial tasks from human intent to restricted quantum hardware remains a severe challenge. To bridge this gap, we present an end-to-end framework integrating a fine-tuned Large Language Model (LLM) front-end with a highly scalable, domain-specific quantum-classical backend. The front-end utilizes Supervised Fine-Tuning (SFT) and Direct Preference Optimization (DPO) to translate free-form natural language instructions into structurally robust Quadratic Unconstrained Binary Optimization (QUBO) constraints without false negatives. To overcome the strict qubit limits of near-term quantum devices, our framework features a novel constraint-preserving graph partitioner and a compressed separator-based dynamic programming (DP) merge. By structurally encoding constraints via W-state initialization and XY-mixers in Conditional Value-at-Risk Quantum Approximate Optimization (CVaR-QAOA), the pipeline stays highly compact. Empirical results demonstrate that this architecture circumvents classical scaling walls, recovering the global optimum on 100\% of idealized oracle cases and 96.3\% under real QAOA sampling, enabling natural-language-guided task allocation at previously intractable scales.
\end{abstract}



\section{Introduction}
\label{sec:intro}
	
    Deploying multi-drone systems for complex spatial tasks, such as regional surveillance, disaster response, and logistics, requires efficient and dynamic zone assignment \cite{zhou2025survey}. Traditionally, human operators must translate high-level mission goals into precise programmatic constraints or mathematical formulations—a process that is time-consuming, error-prone, and requires specialized expertise. With the advent of Large Language Models (LLMs), there is a growing opportunity to bridge this gap by enabling robots to parse free-form, natural-language instructions directly into executable task specifications \cite{yao2023bridging}. However, converting natural language into structured mathematical models is exceptionally challenging, as even minor logical omissions can lead to critical optimization failures.

    To address this initial barrier, recent advances have successfully demonstrated the automated translation of unstructured problem specifications into formal Quadratic Unconstrained Binary Optimization (QUBO) models \cite{dohyun2025a}. Specifically, by leveraging a systematic fine-tuning strategy that integrates Supervised Fine-Tuning (SFT) and Direct Preference Optimization (DPO), LLMs can be optimized to generate structured constraints with high structural accuracy and logical consistency. A key strength of this pipeline is its exceptional robustness in eliminating False Negatives—i.e., missed constraints—thereby ensuring that the critical boundaries of the multi-drone task allocation problem are fully captured from the user's natural language input.

While translating natural language into a QUBO formulation successfully digitizes human intent \cite{glover2018tutorial}, finding the optimal assignment introduces a severe computational bottleneck. As the fleet size and the number of candidate zones increase, the multi-drone assignment problem triggers an exponential combinatorial explosion ($\mathcal{O}(M^N)$), quickly rendering classical exact solvers and exhaustive search methods intractable within real-time operational windows. To shatter this classical barrier, quantum optimization algorithms—specifically the Quantum Approximate Optimization Algorithm (QAOA) \cite{farhi2014quantum}—have emerged as a transformative alternative, offering the potential to explore massive state spaces concurrently. However, directly applying QAOA to large-scale robotics problems is currently blocked by physical hardware limitations. Real-world Noisy Intermediate-Scale Quantum (NISQ) devices are restricted by strict qubit budgets and gate error rates \cite{preskill2018quantum}, meaning a monolithic LLM-generated QUBO instance will quickly overwhelm available quantum capacity. 

To overcome this scalability barrier, this paper presents an end-to-end integration that bridges a robust language front-end \cite{dohyun2025a} with a highly optimized, domain-specific quantum-classical backend. Rather than simply chaining existing computational tools, our primary contribution lies in the hardware-aware formulation and structural optimization tailored specifically to the multi-drone spatial assignment problem. We introduce a domain-specific sub-QUBO encoding that eliminates the severe qubit overhead typically associated with one-hot assignment penalty terms. By mapping the drone-zone constraints directly into W-state initializations and Hamming-weight-preserving XY-mixers within CVaR-QAOA, we create a highly compact representation that structurally enforces hard constraints at the quantum circuit level. Furthermore, we adapt classical graph partitioning and separator-based dynamic programming to the quantum domain, establishing a scalable architecture that decomposes massive, LLM-generated assignment problems into hardware-compatible sub-QUBOs and recombines them while rigorously preserving the global optimum. Through this tailored integration, our framework successfully bypasses the classical exhaustive-search wall, demonstrating a practical blueprint for translating high-level human intent into scalable, globally optimal quantum execution.
 

\section{Related Work}
\label{sec:work}
\textbf{Language-Conditioned Robot Task Planning.} Traditional multi-robot task allocation (MRTA) relies on rigid, hand-coded formal specifications like MILP \cite{khamis2015multi}. While recent frameworks (e.g., SayCan \cite{ahn2022icanisay}, Code as Policies \cite{liang2023code}) enable LLMs to parse natural-language instructions, they mostly target single-robot tasks. Extending LLMs to multi-drone systems via heuristic prompting \cite{kannan2024smart} lacks optimization guarantees and is prone to hallucinated constraints. To bridge this, fine-tuning pipelines (SFT+DPO) have been proposed to reliably emit structurally valid QUBO constraints without false negatives \cite{dohyun2025a}.

\textbf{Quantum Optimization in Robotics.} As drone fleet sizes expand, combinatorial explosion renders classical exact solvers intractable \cite{juan2015review}. QAOA has emerged as a promising quantum alternative for solving NP-hard QUBO problems \cite{farhi2014quantum, blekos2024review, cattelan2024modeling}. However, Noisy Intermediate-Scale Quantum (NISQ) devices are severely constrained by qubit budgets and gate-error rates \cite{preskill2018quantum, lau2022nisq}, making monolithic encoding of large multi-drone instances impossible and necessitating hardware-aware decomposition.

\textbf{Scalable Decomposition.} Classical decomposition methods (e.g., ADMM) incur heavy communication overheads \cite{di2014distributed}. Recent quantum-classical partitioners \cite{verma2022penalty} fit subproblems within NISQ limits but rely on greedy merges that destroy global context. We overcome this using a constraint-preserving partitioner coupled with a compressed separator-DP merge \cite{bodlaender2008combinatorial} that provably recovers the global optimum (Sec.~\ref{sec:dp}).

\section{Problem Formulation}
\label{sec:problem}

Before defining the objective function, we summarize the main symbols used throughout this section in Table~\ref{tab:notation}.

\begin{table}[b]
\centering
\caption{Main symbols used in the problem formulation.}
\label{tab:notation}
\scalebox{0.85}{%
\begin{tabular}{ll}
\toprule
Symbol & Description \\
\midrule
$N$ & Total number of drones \\
$M$ & Total number of candidate zones \\
$\mathcal{D}$ & Set of drones \\
$\mathcal{Z}$ & Set of candidate zones \\
$x_{ij}$ & Binary assignment of drone $i$ to zone $j$ \\
$\mathbf{p}^{\mathrm{d}}_i$ & Position of drone $i$ \\
$\mathbf{p}^{\mathrm{z}}_j$ & Position of zone $j$ \\
$d_{ij}$ & Euclidean travel distance between drone $i$ and zone $j$ \\
$\alpha_i$ & Movement weight for drone $i$ \\
$\mathcal{F}$ & Force assignment set \\
$\mathcal{B}$ & Forbidden assignment set \\
$\rho_j$ & Requested minimum occupancy of zone $j$ \\
$m_j$ & Target occupancy of zone $j$ \\
$n_j$ & Number of drones assigned to zone $j$ \\
$\lambda$ & Occupancy penalty weight \\
$\lambda_{\mathrm{hard}}$ & Hard-constraint penalty weight \\
\bottomrule
\end{tabular}}
\end{table}

We consider a fleet of $N$ drones indexed by 
$i \in \mathcal{D} = \{1, \ldots, N\}$ and 
$M$ candidate zones indexed by 
$j \in \mathcal{Z} = \{1, \ldots, M\}$. 
Drone $i$ and zone $j$ are located at 
$\mathbf{p}^{\mathrm{d}}_i \in \mathbb{R}^2$ and 
$\mathbf{p}^{\mathrm{z}}_j \in \mathbb{R}^2$, respectively. 
We define binary assignment variables 
$x_{ij} \in \{0,1\}$, where $x_{ij}=1$ indicates that drone $i$ is assigned to zone $j$. 
The travel cost between drone $i$ and zone $j$ is given by the Euclidean distance
$d_{ij} = \|\mathbf{p}^{\mathrm{d}}_i - \mathbf{p}^{\mathrm{z}}_j\|_2$.

Each drone must be assigned to exactly one zone, i.e.,
$\sum_{j \in \mathcal{Z}} x_{ij} = 1$ for all $i \in \mathcal{D}$.
This one-hot constraint is enforced structurally by our solver
(Sec.~\ref{sec:method}) rather than through a penalty term.

From the operator instruction, the language front-end extracts:
(i) per-drone movement weights 
$\alpha_i \in \{\alpha_{\mathrm{move}}, 1\}$,
where $\alpha_i = \alpha_{\mathrm{move}} > 1$ indicates that the movement of drone $i$ should be minimized;
(ii) force and forbid assignment sets
$\mathcal{F}, \mathcal{B} \subseteq \mathcal{D} \times \mathcal{Z}$; and
(iii) a requested minimum occupancy $\rho_j \ge 0$ for each zone.

The target occupancy of zone $j$ is defined as
$m_j = \max(L,\rho_j) \ge 1$,
which combines the global minimum coverage level $L$
(default $L=1$) with the zone-specific request $\rho_j$.
Thus, every zone must receive at least one drone unless a larger occupancy target is specified.
Let
$n_j = \sum_i x_{ij}$
denote the number of drones assigned to zone $j$.
The global assignment cost is then formulated as

\begin{equation}
\begin{aligned}
H(\mathbf{x}) = {}&
\sum_{i \in \mathcal{D}}
\sum_{j \in \mathcal{Z}}
\alpha_i d_{ij} x_{ij}
\quad\text{(travel cost)} \\
&+
\lambda_{\mathrm{hard}}
\Bigl(
\sum_{(i,j)\in\mathcal{F}}(1-x_{ij})
+
\sum_{(i,j)\in\mathcal{B}} x_{ij}
\Bigr)
\quad\text{(force/forbid)} \\
&+
\lambda
\sum_{j\in\mathcal{Z}}
\max(0, m_j - n_j)^2
\quad\text{(target occupancy)}.
\end{aligned}
\label{eq:qubo}
\end{equation}

This cost $H$ serves as the evaluation objective:
both the separator-DP merge (Sec.~\ref{sec:dp})
and the brute-force oracle evaluate assignments using the same objective function,
providing a consistent optimality criterion.
Here, $\lambda_{\mathrm{hard}} \gg \lambda$
ensures that any hard-constraint violation dominates feasible costs,
while $\lambda$ controls the occupancy penalty relative to the travel cost.
Although the force/forbid term is written explicitly for completeness,
our implementation enforces these constraints structurally through variable fixing and deletion
(Sec.~\ref{sec:qubo}),
making the corresponding penalty identically zero on feasible one-hot assignments.
The deficit-only occupancy penalty additionally admits a slack-qubit reformulation;
its concrete weighting strategy and the handling of the $m_j \ge 2$ case are deferred to Sec.~\ref{sec:method}.


\section{Method}
\label{sec:method}

  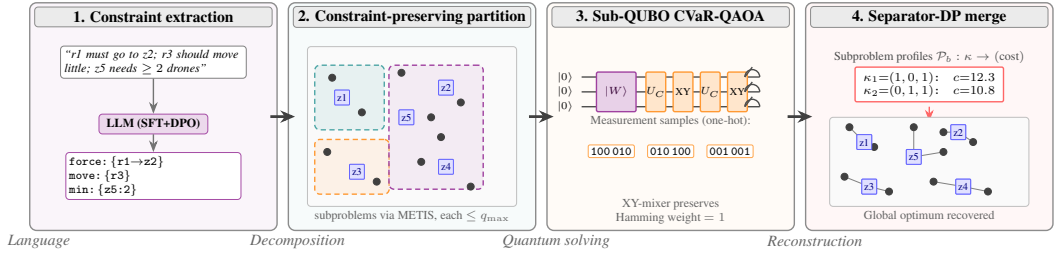
\begin{figure*}[t]
    \centering
    \resizebox{1\textwidth}{!}{\begin{tikzpicture}[
    font=\small,
    >={Stealth[length=2.4mm]},
    panel/.style={
        rounded corners=4pt, draw=black!45, thick,
        minimum width=50mm, minimum height=46mm,
        inner sep=2pt
    },
    paneltitle/.style={font=\footnotesize\bfseries, align=center, draw=black!45,
        fill=black!8, rounded corners=2pt, inner sep=2pt, minimum width=48mm,
        minimum height=4mm},
    catlabel/.style={font=\footnotesize\itshape, text=black!60},
    flow/.style={-{Stealth[length=3mm,width=2.5mm]}, line width=1.6pt, draw=black!75},
    drone/.style={circle, draw=black!75, fill=black!75, inner sep=0pt, minimum size=1.4mm},
    zone/.style={draw=blue!60, fill=blue!12, rounded corners=0.5pt, inner sep=0pt,
        minimum size=3mm, font=\tiny, text=blue!50!black},
    qbit/.style={draw=orange!70, fill=orange!15, minimum size=2mm, inner sep=0pt}
]

\node[panel, fill=violet!4] (p1) {};
\node[paneltitle] at ($(p1.north)+(0,-3mm)$) {1. Constraint extraction};

\node[draw=black!50, fill=white, rounded corners=2pt, inner sep=2.2pt,
      align=left, font=\scriptsize\itshape, text width=36mm]
      at ($(p1.center)+(0,11mm)$) (bubble)
      {``r1 must go to z2; r3 should move little; z5 needs $\ge 2$ drones''};

\node[draw=violet!70, fill=violet!15, rounded corners=2pt, font=\scriptsize\bfseries,
      minimum width=22mm, minimum height=4mm, inner sep=1pt]
      at ($(p1.center)+(0,-1.5mm)$) (llm) {LLM (SFT+DPO)};

\draw[->, thick, draw=black!55] (bubble.south) -- (llm.north);

\node[draw=violet!70, fill=white, rounded corners=2pt, align=left, inner sep=2pt,
      font=\scriptsize\ttfamily, text width=34mm]
      at ($(p1.center)+(0,-12mm)$) (json)
      {force:\,\{r1$\to$z2\}\\ move:\,\{r3\}\\ min:\,\{z5:2\}};

\draw[->, thick, draw=black!55] (llm.south) -- (json.north);

\node[catlabel] at ($(p1.south west)+(2mm,-2mm)$) {Language};

\node[panel, right=2mm of p1, fill=teal!3] (p2) {};
\node[paneltitle] at ($(p2.north)+(0,-3mm)$) {2. Constraint-preserving partition};

\coordinate (mapBL) at ($(p2.south west)+(4mm,5mm)$);
\coordinate (mapTR) at ($(p2.north east)+(-4mm,-9mm)$);
\node[draw=black!25, fill=black!2, fit={(mapBL)(mapTR)}, inner sep=0pt, rounded corners=2pt] (maptile) {};

\node[zone] at ($(mapBL)+(7mm,22mm)$) (zz1) {z1};
\node[zone] at ($(mapBL)+(28mm,24mm)$) (zz2) {z2};
\node[zone] at ($(mapBL)+(10mm,7mm)$) (zz3) {z3};
\node[zone] at ($(mapBL)+(28mm,8mm)$) (zz4) {z4};
\node[zone] at ($(mapBL)+(20mm,18mm)$) (zz5) {z5};

\node[draw=teal!70, densely dashed, thick, rounded corners=3pt, fill=teal!18, fill opacity=0.45,
      fit={($(mapBL)+(2mm,16mm)$) ($(mapBL)+(15mm,28mm)$)}, inner sep=1.5pt] {};
\node[draw=orange!80, densely dashed, thick, rounded corners=3pt, fill=orange!18, fill opacity=0.45,
      fit={($(mapBL)+(2mm,3mm)$) ($(mapBL)+(16mm,13mm)$)}, inner sep=1.5pt] {};
\node[draw=violet!75, densely dashed, thick, rounded corners=3pt, fill=violet!15, fill opacity=0.45,
      fit={($(mapBL)+(17mm,4mm)$) ($(mapBL)+(35mm,28mm)$)}, inner sep=1.5pt] {};

\node[zone] at ($(mapBL)+(7mm,22mm)$) {z1};
\node[zone] at ($(mapBL)+(28mm,24mm)$) {z2};
\node[zone] at ($(mapBL)+(10mm,7mm)$) {z3};
\node[zone] at ($(mapBL)+(28mm,8mm)$) {z4};
\node[zone] at ($(mapBL)+(20mm,18mm)$) {z5};

\foreach \x/\y in {5/26, 11/19, 4/11, 14/5, 19/27, 25/18, 31/21, 27/14, 33/4, 23/9}{
  \node[drone] at ($(mapBL)+(\x mm,\y mm)$) {};
}

\node[font=\scriptsize, text=black!60] at ($(p2.south)+(0,3mm)$) {subproblems via METIS, each $\le q_{\max}$};

\node[catlabel] at ($(p2.south west)+(2mm,-2mm)$) {Decomposition};

\node[panel, right=2mm of p2, fill=orange!3] (p3) {};
\node[paneltitle] at ($(p3.north)+(0,-3mm)$) {3. Sub-QUBO CVaR-QAOA};

\coordinate (q0) at ($(p3.center)+(-18mm,8mm)$);
\foreach \i in {0,1,2}{
  \draw[black!70, thick] ($(q0)+(0,-3mm*\i)$) -- ($(q0)+(36mm,-3mm*\i)$);
  \node[font=\tiny, anchor=east] at ($(q0)+(-0.5mm,-3mm*\i)$) {$|0\rangle$};
}
\draw[draw=violet!75, fill=violet!18, thick, rounded corners=1pt]
      ($(q0)+(3mm,-7mm)$) rectangle ($(q0)+(11mm,1mm)$);
\node[font=\tiny\bfseries, text=violet!40!black] at ($(q0)+(7mm,-3mm)$) {$|W\rangle$};

\foreach \i/\lbl in {0/$U_C$,1/XY,2/$U_C$,3/XY}{
  \pgfmathsetmacro{\xx}{13+\i*5.5}
  \draw[draw=orange!80, fill=orange!20, thick, rounded corners=1pt]
        ($(q0)+(\xx mm,-7mm)$) rectangle ($(q0)+(\xx mm + 4mm,1mm)$);
  \node[font=\tiny] at ($(q0)+(\xx mm + 2mm,-3mm)$) {\lbl};
}
\foreach \i in {0,1,2}{
  \draw[black!70, thick] ($(q0)+(36mm,-3mm*\i)$) arc(0:180:1.5mm);
  \draw[black!70, thick] ($(q0)+(33mm,-3mm*\i)$) -- ($(q0)+(34.5mm,-3mm*\i + 2mm)$);
}

\node[font=\scriptsize, text=black!60] at ($(p3.center)+(0,-1mm)$) {Measurement samples (one-hot):};

\node[font=\scriptsize\ttfamily, draw=orange!70, fill=white, rounded corners=1pt, inner sep=1.5pt]
     at ($(p3.center)+(-12mm,-7mm)$) {100\,010};
\node[font=\scriptsize\ttfamily, draw=orange!70, fill=white, rounded corners=1pt, inner sep=1.5pt]
     at ($(p3.center)+(0mm,-7mm)$) {010\,100};
\node[font=\scriptsize\ttfamily, draw=orange!70, fill=white, rounded corners=1pt, inner sep=1.5pt]
     at ($(p3.center)+(12mm,-7mm)$) {001\,001};

\node[font=\scriptsize, text=black!60, align=center] at ($(p3.south)+(0,4mm)$)
     {XY-mixer preserves\\ Hamming weight $=1$};

\node[catlabel] at ($(p3.south west)+(2mm,-2mm)$) {Quantum solving};

\node[panel, right=2mm of p3, fill=red!3] (p4) {};
\node[paneltitle] at ($(p4.north)+(0,-3mm)$) {4. Separator-DP merge};

\node[font=\scriptsize, text=black!70] at ($(p4.center)+(0,12mm)$) {Subproblem profiles $\mathcal{P}_b: \kappa\to(\text{cost})$};
\draw[draw=red!60, fill=white, rounded corners=1pt, thick]
     ($(p4.center)+(-15mm,3mm)$) rectangle ($(p4.center)+(15mm,10mm)$);
\node[font=\scriptsize\ttfamily, align=left] at ($(p4.center)+(0,6.5mm)$)
     {$\kappa_1{=}(1,0,1)$: $c{=}12.3$\\ $\kappa_2{=}(0,1,1)$: $c{=}10.8$};

\draw[->, thick, draw=red!60] ($(p4.center)+(0,2mm)$) -- ($(p4.center)+(0,-1mm)$);

\coordinate (fmBL) at ($(p4.south west)+(5mm,5mm)$);
\coordinate (fmTR) at ($(p4.south east)+(-5mm,5mm)+(0,18mm)$);
\node[draw=black!25, fill=black!2, fit={(fmBL)(fmTR)}, inner sep=0pt, rounded corners=2pt] (fmap) {};

\node[zone] at ($(fmBL)+(7mm,13mm)$) (fz1) {z1};
\node[zone] at ($(fmBL)+(26mm,15mm)$) (fz2) {z2};
\node[zone] at ($(fmBL)+(8mm,4mm)$) (fz3) {z3};
\node[zone] at ($(fmBL)+(27mm,4mm)$) (fz4) {z4};
\node[zone] at ($(fmBL)+(17mm,10mm)$) (fz5) {z5};

\foreach \dx/\dy/\zx/\zy in {
    4/16/7/13,
    9/12/7/13,
    3/6/8/4,
    12/3/8/4,
    17/16/17/10,
    23/11/17/10,
    23/15/26/15,
    29/13/26/15,
    32/3/27/4,
    21/5/27/4
}{
  \draw[black!50, thin] ($(fmBL)+(\dx mm,\dy mm)$) -- ($(fmBL)+(\zx mm,\zy mm)$);
  \node[drone] at ($(fmBL)+(\dx mm,\dy mm)$) {};
}
\node[zone] at ($(fmBL)+(7mm,13mm)$) {z1};
\node[zone] at ($(fmBL)+(26mm,15mm)$) {z2};
\node[zone] at ($(fmBL)+(8mm,4mm)$) {z3};
\node[zone] at ($(fmBL)+(27mm,4mm)$) {z4};
\node[zone] at ($(fmBL)+(17mm,10mm)$) {z5};

\node[font=\scriptsize, text=black!60] at ($(p4.south)+(0,3mm)$) {Global optimum recovered};

\node[catlabel] at ($(p4.south west)+(2mm,-2mm)$) {Reconstruction};

\draw[flow, shorten >=-1.5mm, shorten <=-1.5mm] (p1.east) -- (p2.west);
\draw[flow, shorten >=-1.5mm, shorten <=-1.5mm] (p2.east) -- (p3.west);
\draw[flow, shorten >=-1.5mm, shorten <=-1.5mm] (p3.east) -- (p4.west);

\end{tikzpicture}}
    \caption{End-to-end pipeline: an instruction is parsed by the fine-tuned LLM (SFT+DPO) into constraints; the constraint-aware graph is partitioned into qubit-budget subproblems; each subproblem is solved by CVaR-QAOA (W-state init + one-hot-preserving XY-mixer); per-subproblem profiles are recombined by a separator-DP merge.}
    \label{fig:pipeline}
  \end{figure*}

The pipeline comprises four stages (Fig.~\ref{fig:pipeline}): natural-language constraint extraction, constraint-preserving partitioning, sub-QUBO CVaR-QAOA optimization, and separator-DP merge.

\subsection{Natural-Language Constraint Extraction}
\label{sec:llm}
The entry point is a free-form instruction. We employ a fine-tuned LLM (Qwen1.5-1.8B~\citep{bai2023qwen} with LoRA~\citep{hu2022lora}) to translate instructions into structured constraints (force/forbid assignments, minimum coverage). Supervised Fine-Tuning (SFT)~\cite{ouyang2022training} fixes the output schema, while Direct Preference Optimization (DPO)~\cite{rafailov2023direct} maximizes structural validity and essentially eliminates missing constraints. A deterministic layer then validates these extracted constraints and emits the global QUBO formulation.

\subsection{Constraint-Preserving Graph Partitioning}
\label{sec:partition}
To overcome monolithic qubit limits, we partition the drone-zone assignment graph into qubit-budgeted subproblems using PyMETIS~\citep{karypis1998fast}. At this stage, our specific structural contribution is the introduction of a coverage candidate pruning mechanism prior to graph partitioning. Crucially, to mathematically preserve the global optimum across partitions, any drone whose detour to a zone exceeds the coverage penalty threshold ($\lambda L^2$) is excluded. This targeted pruning safely removes assignments that cannot improve the global objective, ensuring that the heuristic partitioning does not inadvertently sever the globally optimal solution set. Subproblems are strictly sized to fit the maximum hardware qubit budget ($q_{\max}$).

\subsection{Sub-QUBO Encoding and CVaR-QAOA}
\label{sec:qubo}
We structurally reduce the qubit footprint before quantum execution. First, through \emph{variable fixing and deletion}, force/forbid constraints directly eliminate their respective decision qubits without requiring penalty terms. 

 For the remaining variables, our core contribution at the circuit level is a domain-specific structural encoding that bypasses the heavy qubit overhead of traditional one-hot penalty terms. We structurally group each drone's decision qubits into a Hamming-weight-one block of size k. Instead of a generic transverse-field mixer, we initialize this block in the W-state via a cascade of controlled-$R_y$ and CNOT gates~\citep{cruz2019efficient}, creating an equal superposition of all valid single-zone assignments: $|W\rangle = \frac{1}{\sqrt{k}}\bigl(|10\cdots0\rangle + |01\cdots0\rangle + \cdots + |0\cdots01\rangle\bigr)$. We then apply a block-local XY-mixer~\citep{hadfield2019quantum}, $\sum_{a<b} (X_a X_b + Y_a Y_b)$, which preserves the Hamming weight of each block throughout the quantum evolution. By restricting the quantum evolution exclusively to the valid one-hot subspace, this bespoke encoding fundamentally improves the convergence landscape for the CVaR-QAOA solver.

Each sub-QUBO is optimized via CVaR-QAOA~\cite{barkoutsos2020improving} using the COBYLA optimizer~\citep{powell1994direct}. For zones requiring multiple drones ($m_j \geq 2$), we employ an exact binary slack expansion, while single-drone coverage is delegated to the classical merge layer to save qubits.

\subsection{Compressed Separator-DP Merge}
\label{sec:dp}
The subproblem profiles are recombined using a left-to-right dynamic program (DP) over the partition separators. While standard DP recombination quickly suffers from combinatorial explosion on dense graphs, our algorithmic contribution here is the design of a lossless compressed boundary state. By clamping zone counts strictly at their target occupancy $m_j$, the DP footprint is drastically reduced. This compression enables an exact dominance pruning strategy: among partial assignments yielding the same clamped boundary state, only the one with the minimum accumulated travel cost is retained. Because the final coverage penalty depends solely on these clamped counts (separability and sufficiency), we establish a formal guarantee that this highly compressed search definitively recovers the global optimum of the original monolithic QUBO. 


\section{Experiments}
\label{sec:experiments}

The primary objective of our evaluation is to empirically validate the core contributions introduced in Section~\ref{sec:intro}. Specifically, we aim to demonstrate that (1) the structurally encoded CVaR-QAOA can effectively optimize the sub-QUBOs within hardware constraints, and (2) the partition-and-merge architecture bypasses classical scaling bottlenecks while preserving the global optimum.

\textbf{Scenarios \& Hardware.} 
We evaluate $300$ scenarios across $10$ fleet configurations, denoted as $N{\times}M$ (number of drones $\times$ number of zones), ranging from $7{\times}5$ to $12{\times}4$ on a $50{\times}50$ map. Each scenario includes random force/forbid constraints ($\le 3$), movement preferences on ${\sim}20\%$ of drones ($\alpha_{\mathrm{move}}{=}3$, otherwise $1$), and minimum zone requests on ${\sim}25\%$ of zones, with global coverage strictly enforced. Experiments and brute-force baselines run on a single NVIDIA RTX~4080 using Qiskit-Aer~\citep{javadi2024quantum} noiseless statevector simulation. Penalty weights are set to $\lambda = 5\bar{d}$ (where $\bar{d}$ is the mean drone--zone distance) and $\lambda_{\mathrm{hard}} = 1000\lambda$.

\textbf{Hyperparameters.}
Subproblems are partitioned with a maximum qubit budget $q_{\max}{=}24$. To reduce unnecessary candidate assignments, the coverage candidate pruning described in Sec.~\ref{sec:partition} is applied prior to partitioning, recursively bisecting subproblems if the qubit budget is exceeded. Each sub-QUBO is optimized via CVaR-QAOA ($p{=}6$, $\alpha{=}0.1$) using COBYLA with $8$ restarts ($1$ INTERP~\citep{zhou2020quantum}, $7$ random). The shot count and iteration budget are scaled linearly with the sub-QUBO size to maintain consistent optimization quality across different partition sizes.

 \subsection{Constraint-Extraction Fidelity}
  \label{sec:llm-eval}

  \begin{table}[b]
    \centering
    \caption{Constraint-extraction fidelity on $720$ gold constraints (micro-averaged).
    FN $=$ missed constraints (lower is better).}
    \label{tab:llm}
    \scalebox{0.85}{%
    \begin{tabular}{lcccc}
      \toprule
      Method & Precision & Recall & F1 & FN \\
      \midrule
      spaCy-based      & 0.406 & 0.524 & 0.458 & 340 \\
      SRL-based        & 0.603 & 0.436 & 0.506 & 403 \\
      OpenIE-based     & 0.465 & 0.699 & 0.559 & 215 \\
      LLM (SFT-only)   & 0.947 & 0.869 & 0.907 & 94 \\
      \textbf{LLM+DPO (Ours)} & \textbf{0.984} & \textbf{0.997} & \textbf{0.990} & \textbf{2} \\
      \bottomrule
    \end{tabular}}
  \end{table}
  
  Before optimization, the language front-end (Sec.~\ref{sec:llm}) must translate the
  operator instruction into the correct constraint set: a single missed constraint
  yields an invalid QUBO and a meaningless assignment. We therefore evaluate
  constraint-extraction fidelity on a benchmark of $720$ gold constraints spanning
  three difficulty tiers---C1 (single-clause), C2 (medium), and C345
  (multi-clause/complex)---scoring each predicted constraint against the gold JSON.
  We compare three rule-based NLP extractors (spaCy, SRL, OpenIE), an SFT-only
  ablation, and our SFT+DPO front-end.

  Table~\ref{tab:llm} reports micro-averaged precision, recall, and F1, together with
  the absolute number of missed constraints (FN). Rule-based extractors plateau at
  $\text{F1}\leq 0.56$, as fixed patterns fail to generalize across phrasings. The
  SFT-only model reaches F1 $0.91$ but still misses $94$ constraints. Adding DPO drives
  recall to $0.997$ and cuts missed constraints to $2$ of $720$---an essentially
  false-negative-free front-end, the property the downstream optimization depends on:
  every extracted boundary is preserved, so no hard constraint is silently dropped
  before the QUBO is built.

To illustrate this automated pipeline, consider a simple scenario with two drones (r1, r2) and two zones (z1, z2). An operator issues the natural language instruction: \textit{``r1 must not be assigned to zone z2, and Zone z1 must be assigned at least 2 drones.''} The SFT+DPO front-end parses this text and successfully extracts two structured constraints: a forbid condition $\mathcal{B} = \{(1,2)\}$ and a minimum occupancy target $\rho_1 = 2$. Consequently, our deterministic layer automatically compiles these parameters into the global QUBO formulation (following Eq.~\ref{eq:qubo}):
\begin{equation}
H(\mathbf{x}) = \underbrace{\sum_{i=1}^{2}\sum_{j=1}^{2} d_{ij} x_{ij}}_{\text{Travel Cost}} + \underbrace{\lambda_{\mathrm{hard}} x_{12}}_{\text{Forbid } (r_1 \to z_2)} +
\underbrace{\lambda \max(0, 2 - (x_{11} + x_{21}))^2}_{\text{Coverage } (z_1 \ge 2)}.
\label{eq:qubo_example}
\end{equation}
This empirical translation confirms that our front-end flawlessly captures both boundary limitations and regional demands from free-form operator intent.

\subsection{Global-Optimum Attainment}

We evaluate the pipeline along two complementary axes. First, to isolate the
correctness of the partition-and-merge layer from QAOA's probabilistic sampling,
we replace the QAOA solver with an exhaustive subproblem enumerator: any deviation from the brute-force global optimum must then be
attributed to the partition or merge layer. Second, we run the full pipeline with
real CVaR-QAOA.

\begin{table}[b]
  \centering
  \caption{Global-optimum attainment of the pipeline, with an idealized
  subproblem oracle and with real CVaR-QAOA on a single RTX~4080 GPU.
  }
  \label{tab:goal1}
  \scalebox{0.85}{%
  \begin{tabular}{lcc cc ccc}
    \toprule
    \multirow{2}{*}{Config} & \multirow{2}{*}{$N/M$} & \multirow{2}{*}{$n$}
      & \multicolumn{2}{c}{Oracle}
      & \multicolumn{3}{c}{Real CVaR-QAOA} \\
    \cmidrule(lr){4-5}\cmidrule(lr){6-8}
      & & & Match & Acc.\ (\%) & Match & Acc.\ (\%) & Avg.\ gap (\%) \\
    \midrule
    $7\times 5$  & 1.4 & 30 & 30 & 100.0 & 30 & 100.0 & 0.00 \\
    $8\times 4$  & 2.0 & 30 & 30 & 100.0 & 28 & 93.3 & 0.05 \\
    $8\times 5$  & 1.6 & 30 & 30 & 100.0 & 29 & 96.7 & 0.05 \\
    $9\times 4$  & 2.2 & 30 & 30 & 100.0 & 29 & 96.7 & 0.01 \\
    $9\times 5$  & 1.8 & 30 & 30 & 100.0 & 30 & 100.0 & 0.00 \\
    $10\times 4$ & 2.5 & 30 & 30 & 100.0 & 28 & 93.3 & 0.07 \\
    $10\times 5$ & 2.0 & 30 & 30 & 100.0 & 30 & 100.0 & 0.00 \\
    $11\times 4$ & 2.8 & 30 & 30 & 100.0 & 27 & 90.0 & 0.13 \\
    $11\times 5$ & 2.2 & 30 & 30 & 100.0 & 28 & 93.3 & 0.07 \\
    $12\times 4$ & 3.0 & 30 & 30 & 100.0 & 30 & 100.0 & 0.00 \\
    \midrule
    \textbf{Total} & & \textbf{300}
      & \textbf{300} & \textbf{100.0}
      & \textbf{289} & \textbf{96.3} & \textbf{0.04} \\
    \bottomrule
  \end{tabular}}
\end{table}

Table~\ref{tab:goal1} reports both columns side by side. The brute-force oracle column recovers the global optimum on 300/300 (100.0\%) scenarios across all
configurations, empirically confirming the optimality guarantee of
Sec.~\ref{sec:dp} and the conservative completeness of the constraint-preserving
partition. Under real CVaR-QAOA, the pipeline attains the exact global optimum in \textbf{289/300 cases (96.3\%)}, with a negligible average optimality gap of 0.04\% across all 300 scenarios. To contextualize this achievement, finding the absolute global minimum in a discrete state space of O($M^N$) is exceptionally difficult for heuristic solvers, which frequently get trapped in local minima. Achieving a 96.3\% exact-match rate is remarkably high for a probabilistic quantum solver operating at a fixed shot count. This high attainment rate directly validates the efficacy of our structural sub-QUBO encoding: by utilizing W-state initialization and XY-mixers to enforce one-hot constraints, the quantum solver avoids the optimization landscape distortions (e.g., barren plateaus) typically caused by heavy penalty terms, allowing it to efficiently locate the optimum within a strict qubit budget. Every sub-optimal case is therefore attributable to incomplete QAOA sampling at the configured shot count rather than structural flaws; rerunning these sub-optimal scenarios recovers the global optimum, reflecting QAOA's inherent stochasticity.

\subsection{Scalability and Runtime Comparison}

  \begin{figure}[b]
    \centering
    \input{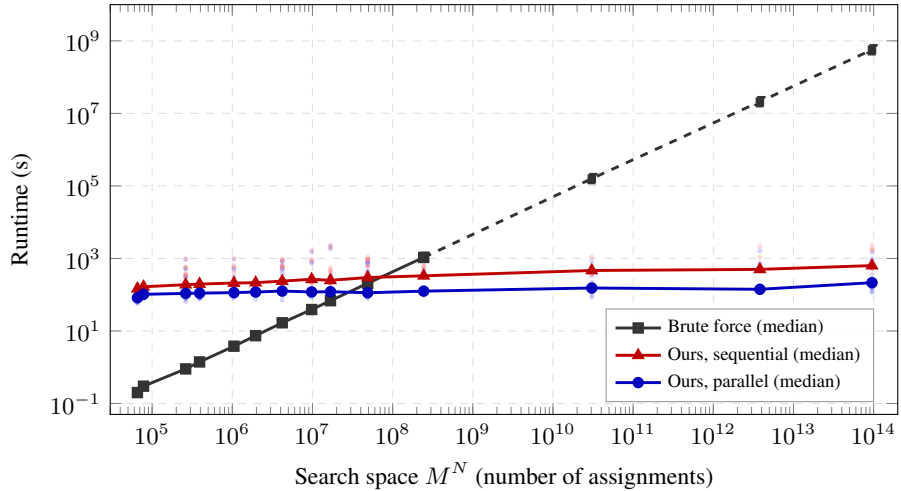}
    \caption{Runtime vs.\ problem scale. Brute-force scales as $M^N$ (measured;
    dashed extrapolation), while the pipeline is bounded by the slowest sub-QUBO's QAOA
    plus a polynomial merge; the parallel variant (blue) runs sub-QUBOs concurrently.
    Past the crossover ($M^N \approx 5 \times 10^7$) brute force becomes intractable.}
    \label{fig:runtime}
  \end{figure}
  
Figure~\ref{fig:runtime} compares the median wall-clock runtime as the global problem scale increases. To thoroughly evaluate scalability, we extended the scenarios up to a $20{\times}5$ (drones ${\times}$ zones) configuration. 

Classical brute-force exhaustive search evaluates every valid one-hot assignment, resulting in a time complexity of $\mathcal{O}(M^N)$. We directly measured brute-force runtimes up to the $12{\times}5$ configuration. For the larger, intractable scales ($15{\times}5$, $18{\times}5$, and $20{\times}5$), the brute-force runtimes were theoretically extrapolated based on this $\mathcal{O}(M^N)$ complexity using an atomic evaluation function benchmark. As depicted, the exhaustive approach rapidly hits an exponential wall.

In contrast, our partition-and-merge architecture successfully avoids this exponential explosion. Validating our second core contribution, the runtime of our pipeline is dominated by the slowest sub-QUBO's QAOA execution, which remains strictly bounded by the constant qubit budget $q_{\max}$. Because the structural encoding keeps each subproblem compact, and the separator-DP merge adds only polynomial overhead, the framework scales gracefully. We further accelerate the pipeline via a parallel variant that allocates one GPU stream per subproblem. Rather than a fixed constant speedup, the efficiency of this parallelization dynamically depends on the number of partitions and the variance in subproblem sizes. As the global problem expands and generates more subproblems, this concurrent execution naturally absorbs the scaling burden, making the parallel advantage increasingly pronounced. The crossover point where our parallel pipeline overtakes the brute-force approach occurs near a state space of $M^N \approx 5 \times 10^7$, beyond which the runtime gap widens by orders of magnitude in our framework's favor.

To formally quantify this profound trade-off between exactness and scalability, we introduce a Time-Normalized Effective Accuracy metric, defined as the probability of finding the global optimum divided by the execution time ($\frac{\text{Attainment Rate}}{\text{Runtime}}$). For small-scale configurations (e.g., $7{\times}5$), the brute-force oracle evaluates almost instantly, yielding a high effective accuracy. However, as the scale approaches the crossover point ($15{\times}5$ and beyond), the brute-force runtime explodes into hours or days, causing its effective accuracy to asymptotically collapse toward zero despite its $100\%$ attainment rate. Conversely, our quantum-classical pipeline sustains a highly stable $96.3\%$ global-optimum attainment rate while maintaining a nearly flat, sub-minute runtime trajectory bounded by $q_{\max}$. Consequently, past the crossover threshold, our framework achieves an effective accuracy that is orders of magnitude superior to exhaustive search, decisively proving that trading a negligible $3.7\%$ probabilistic gap for an exponential speedup is highly advantageous for real-time fleet coordination.
\section{Limitations}
\label{sec:limitations}
While the proposed framework successfully scales multi-drone task allocation, we note two practical considerations for future deployment. 

\textbf{Physical NISQ Deployment.} Our experiments utilize noiseless statevector simulations to validate the fundamental architecture of the decomposition method. Transitioning to physical quantum hardware will naturally introduce gate noise and amplify the inherent probabilistic sampling variance of CVaR-QAOA. Maintaining subproblem optimality in real-world deployments will thus require coupling our pipeline with robust error mitigation and adaptive sampling strategies.

\textbf{Decomposition Trade-offs.} The classical overhead of the separator-DP merge is tied to the graph's separator size. Shrinking partitions to fit extremely tight qubit budgets inevitably increases this boundary state, potentially bottlenecking the classical recombination phase for densely constrained topologies. Careful calibration of the partition size is required to balance quantum execution with classical merge efficiency.

\section{Conclusion}
\label{sec:conclusion}
We presented an end-to-end framework that bridges the gap between natural-language instructions and scalable quantum optimization for multi-drone task allocation. Rather than heuristically assembling off-the-shelf solvers, we demonstrated a domain-specific quantum-classical backend tailored to bypass the strict qubit limitations of NISQ hardware. By structurally encoding hard assignment constraints via W-state initialization and XY-mixers, we eliminated severe penalty overheads, allowing sub-QUBOs to remain highly compact and executable. Furthermore, our application of constraint-preserving graph partitioning and compressed separator-DP merge ensured that these manageable subproblems could be recombined without sacrificing the global optimum.

Empirically, our pipeline validates this tailored integration. The SFT+DPO language front-end captures structural boundaries with $99.7\%$ recall. Passed to the quantum backend, our structural encoding and merge layer recovered the global optimum in $100\%$ of oracle evaluations and $96.3\%$ of cases using real CVaR-QAOA sampling, maintaining a sub-percent optimality gap. Most importantly, this architecture successfully circumvents the classical exhaustive-search bottleneck, scaling gracefully as the global fleet size expands.

Future work will focus on transitioning from statevector simulations to physical NISQ devices, which will require coupling our structural encoding with hardware-aware error mitigation and adaptive sampling loops. Ultimately, we aim to explore a Hierarchical QAOA (QAOA-in-QAOA) structure, replacing the classical DP merge with a secondary quantum optimization layer to establish a fully quantum-centric scaling path for massive autonomous fleets.

\clearpage


\bibliography{reference}  

\end{document}